\documentclass[10pt,twocolumn,letterpaper]{article}

\usepackage{iccv}
\usepackage{times}
\usepackage{epsfig}
\usepackage{graphicx}
\usepackage{amsmath}
\usepackage{amssymb}

\usepackage[pagebackref=true,breaklinks=true,letterpaper=true,colorlinks,bookmarks=false]{hyperref}

\usepackage{booktabs}
\usepackage{multirow}
\usepackage{caption}
\usepackage{subcaption}
\usepackage{float}
\usepackage{subcaption}
\usepackage{enumitem}
\usepackage{numprint}
\usepackage{xcolor}
\usepackage{tabu}
\usepackage[numbers]{natbib}
\usepackage{textcomp}
\usepackage{pythonhighlight}

\definecolor{GreenCanvas}{HTML}{F1F8EC}
\definecolor{RedCanvas}{HTML}{FCE9DC}

\usepackage[capitalize]{cleveref}
\crefname{section}{Sec.}{Secs.}
\Crefname{section}{Section}{Sections}
\Crefname{table}{Table}{Tables}
\crefname{table}{Tab.}{Tabs.}

\def\iccvPaperID{4677} % *** Enter the ICCV Paper ID here

\iccvfinalcopy % *** Uncomment this line for the final submission

\ificcvfinal\fi

\newcommand{\ARCH}{Text2Tex\xspace}
\newcommand{\TITLE}{\ARCH: Text-driven Texture Synthesis via Diffusion Models}

\newcommand{\mypara}[1]{\noindent\textbf{#1}}

\ificcvfinal\fi

\begin{document}

%%%%%%%%% TITLE
\title{\TITLE}

\author{
Dave Zhenyu Chen$^{1}$ \quad Yawar Siddiqui$^{1}$ \quad Hsin-Ying Lee$^{2}$ \quad Sergey Tulyakov$^{2}$ \quad Matthias Nie{\ss}ner$^{1}$\\
$^{1}$Technical University of Munich \qquad $^{2}$Snap Research \\
\url{https://daveredrum.github.io/Text2Tex/} \\
}

% \maketitle

\twocolumn[{%
	\renewcommand\twocolumn[1][]{#1}%
	\maketitle
	\thispagestyle{empty}
	\begin{center}
		\includegraphics[width=\textwidth]{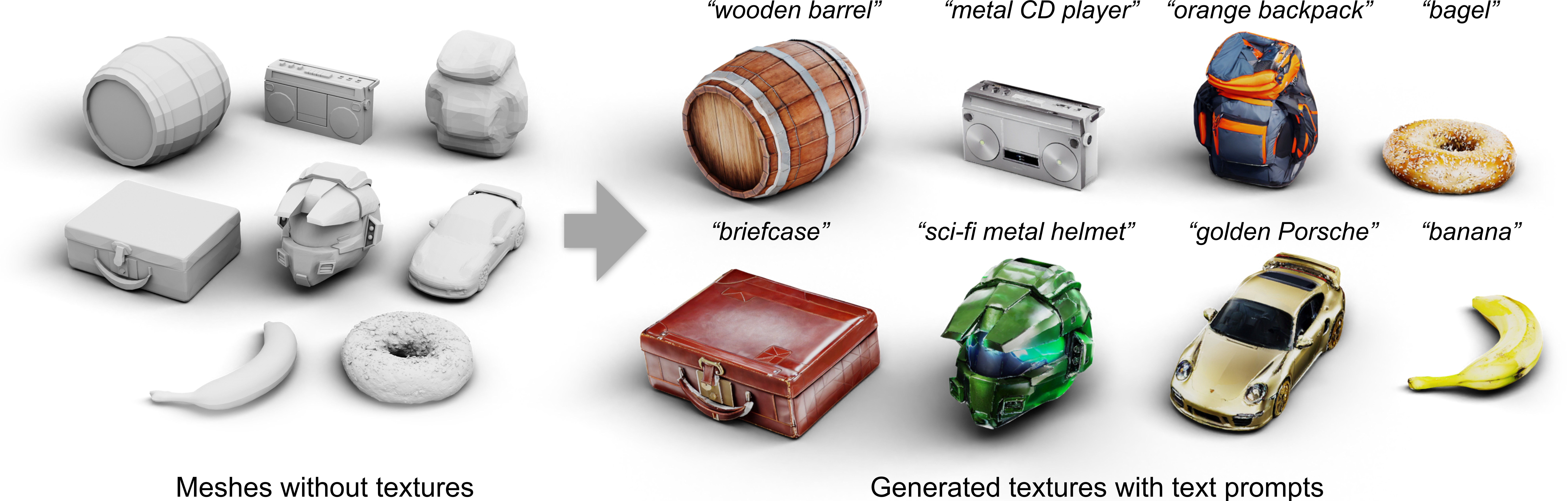}
		\captionof{figure}{
		We introduce~\ARCH, a text-driven architecture for 3D texture synthesis. Given object geometries and text prompts as input, ~\ARCH generates high quality and consistent textures via depth-aware inpainting and refinement.
		}
		\label{fig:teaser}
	\end{center}
}]

\ificcvfinal\thispagestyle{empty}\fi

%%%%%%%%% ABSTRACT
\begin{abstract}

We present Text2Tex, a novel method for generating high-quality textures for 3D meshes from the given text prompts.
Our method incorporates inpainting into a pre-trained depth-aware image diffusion model to progressively synthesize high resolution partial textures from multiple viewpoints. 
To avoid accumulating inconsistent and stretched artifacts across views, we dynamically segment the rendered view into a generation mask, which represents the generation status of each visible texel. 
This partitioned view representation guides the depth-aware inpainting model to generate and update partial textures for the corresponding regions.
Furthermore, we propose an automatic view sequence generation scheme to determine the next best view for updating the partial texture.
Extensive experiments demonstrate that our method significantly outperforms the existing text-driven approaches and GAN-based methods.

\vspace{-3mm}

\end{abstract}

%%%%%%%%% BODY TEXT

\section{Introduction}

Generating high-quality 3D content is an essential component of visual applications in films, games, and upcoming AR/VR scenarios. 
With an increasing number of 3D content datasets, the computer vision community has witnessed significant progress in the field of 3D geometry generation~\cite{chen2019learning, nash2020polygen, wu2018learning, zeng2022lion, muller2022diffrf}. 
Despite the remarkable success in modeling 3D geometries in recent years, fully automatic 3D content generation is still hindered by the laborious human efforts required to design textures. 
Therefore, automating the texture design process through alternative guidance, such as text, has become an intriguing but challenging research problem.

Recently, text-to-image generators have shown remarkable progress in the 2D domain leveraging diffusion model architectures, enabling high resolution 2D content generation based on textual descriptions~\cite{stablediffusion, ramesh2022hierarchical}.
However, there are notable challenges for producing 3D textures via such 2D vision-language prior knowledge.
Specifically, the synthesized textures are expected to be not only with high fidelity to the language cues, but also of high and consistent quality for target meshes.
As such, previous attempts to paint 3D geometry from text inputs often fail to deliver well-textured 3D content.

In this paper, we introduce~\textbf{\ARCH}, a novel texture synthesis method that seamlessly texturizes 3D objects using a pre-trained depth-aware text-to-image diffusion model.
The method renders a target mesh from multiple viewpoints and inpaints the missing appearance with a depth-aware text-to-image diffusion model.
\ARCH follows a \textit{generate-then-refine} strategy. 
Our method progressively generates partial textures across viewpoints and back-projects them to texture space. 
To address stretched and inconsistent artifacts observed from rotated viewpoints, we design a \textit{view partitioning technique} that computes similarity maps between visible texel's normal vectors and the current view direction. 
The generation mask created from these similarity maps guides the diffusion process by indicating regions to generate, update, keep, or ignore. 
This allows us to apply different diffusion strengths to respective regions, inpainting missing appearance and updating stretched artifacts.
However, the autoregressive generation process via the diffusion-based image inpainting model presents a new challenge. 
As the inpainting and updating scheme is conditioned on previously synthesized results, a viewpoint sequence with an ill-defined order or incomplete coverage over the mesh surface may result in unsatisfactory texturization. 
Therefore, we propose an \textit{automatic viewpoint selection} technique that progressively selects the next best view. 
The confidence of each candidate view containing the biggest relative area for generation and updating is estimated, given the partially textured mesh. 
This approach ensures complete coverage over the mesh surface and a high-quality texture map by consistently updating stretched regions.

We demonstrate the effectiveness of \ARCH for synthesizing high-quality 3D textures from language cues. 
The proposed method performs favorably against other language-based texture synthesis methods in terms of FID~\cite{heusel2017gans}, KID~\cite{binkowski2018demystifying}, and user study on a subset of the Objaverse dataset~\cite{objaverse}.
Additionally, our method also outperforms category-specific GAN-based methods on the ShapeNet car dataset~\cite{chang2015shapenet}. 

To summarize, our technical contributions are threefold:

\begin{itemize}

\item We design a novel method for high-quality texture synthesis by progressively inpainting and updating the 3D textures via depth-aware diffusion models.
\item We propose an automatic view sequence generation scheme to dynamically determine the order for generating and updating the texture space.
\item We conduct extensive study on a considerable amount of 3D objects, demonstrating the proposed method is effective for large-scale 3D content generation.

\end{itemize}

\section{Related work}

\mypara{3D Generation from 3D and 2D data.}
To achieve 3D generation, it is natural to train models directly on 3D data. 
In contrast to 2D images, there are several 3D representations available, each with its unique characteristics, leading to the development of various generative models such as those based on voxels~\cite{lin2023infinicity,siarohin2023unsupervised,smith2017improved,xie2018learning}, point clouds~\cite{achlioptas2018learning,luo2021diffusion}, meshes~\cite{zhang2021sketch2model}, signed distance function~\cite{chen2019learning,cheng2022sdfusion,cheng2022cross,dai2021spsg,autosdf2022}, etc.
However, unlike images or videos that are ubiquitous, 3D data is inherently scarce and challenging to collect and annotate. 
Consequently, the synthesized samples from 3D generative models, trained on 3D data, are of limited quality and diversity, in terms of both structure and texture. Recent works have leveraged differentiable rendering to learn texture generation using only 2D images~\cite{gao2022get3d,siddiqui2022texturify, yu2021learning}. However, they are typically trained for specific shape categories and struggle in the quality of textures.    

\mypara{Text-Guided Generation.}
Recently, there has been tremendous progress in the vision-language domain~\cite{hu2021unit, singh2022flava, wang2022ofa, radford2021learning, chen2022d3net, chen2020scanrefer, chen2021scan2cap, chen2022unit3d}. Specifically, the emergence of Contrastive Language-Image Pre-Training (CLIP)~\cite{radford2021learning} has enabled the development of text-guided image generation through its semantically rich representation trained on text-image pairs. 
Initial efforts~\cite{crowson2022vqgan,patashnik2021styleclip} incorporated CLIP with different backbones, such as StyleGAN~\cite{karras2020analyzing} and VQGAN~\cite{esser2021taming}. 
However, diffusion models~\cite{dhariwal2021diffusion,ho2020denoising,ho2021cascaded,nichol2021improved,saharia2022image}, which have gained attention due to their superior visual quality and training stability compared to Generative Adversarial Networks (GANs)~\cite{GANs}, have recently been trained on large-scale text-image datasets with CLIP encodings~\cite{kawar2022imagic,ramesh2021zero,Rombach_2022_CVPR}. 
Among these models, Stable Diffusion~\cite{stablediffusion,Rombach_2022_CVPR} has garnered significant interest as an open-sourced model with numerous extensions~\cite{stablediffusion,zhang2023adding} that support different conditional modalities in addition to text prompts, including depth images, poses, sketches, etc. 
Additionally, CLIP has also been adopted in 3D to perform text-guided shape and texture generation~\cite{michel2022text2mesh,mohammad2022clip}. 
In this work, we take advantage of the depth-conditioning feature of Stable Diffusion to provide more consistent texturing.

\begin{figure*}[!ht]
    \centering
    \includegraphics[width=0.99\linewidth]{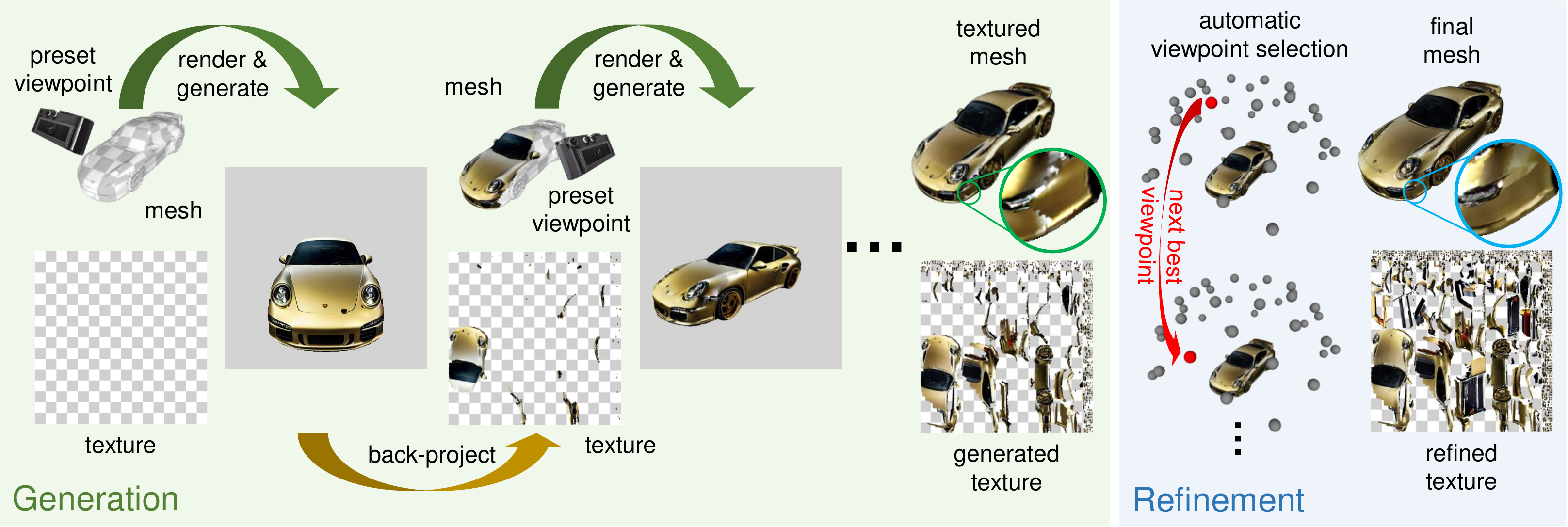}
    \caption{ 
    Overview of \ARCH.
    We illustrate the pipeline using a 3D car mesh with a prompt ``\textit{golden Porsche}".
    We progressively generate the texture via a \textit{generate-then-refine} scheme. 
    In \textbf{progressive texture generation} (Sec.~\ref{sec: generation}), we start by rendering the object from an initial preset viewpoint. We generate a new appearance according to the input prompt via a depth-to-image diffusion model, and project the generated image back to the partial texture. Then, we repeat this process until the last preset viewpoint to output the initial textured mesh. 
    In the subsequent \textbf{texture refinement } (Sec.~\ref{sec: refinement}), we update the initial texture from a sequence of automatically selected viewpoints to refine the stretched and blurry artifacts.
    }
    \label{fig:progressive}
\end{figure*}

\mypara{Text-to-3D from 2D data.}
Inspired by the success of Neural Radiance Fields (NeRF)~\cite{mildenhall2020nerf}, NeRF-based generators~\cite{abdal20233davatargan, EG3D,piGAN,  StyleNeRF,Giraffe, or2022stylesdf,GRAF,skorokhodov3d,xu2022discoscene} have been proposed to learn 3D structures from 2D images using GAN-based frameworks. 
A new research direction emerged by combining NeRF techniques with flourishing diffusion-based text-to-image models, enabling text-to-3D learning with only 2D supervision. 
To address the challenge of optimizing a NeRF field, a score-distillation loss is proposed~\cite{poole2022dreamfusion} that leverages a pretrained 2D diffusion model as a critic to provide essential gradients. 
Subsequent efforts have focused on adopting this loss in latent space~\cite{cheng2022sdfusion,metzer2022latent} and in a coarse-to-fine refinement approach~\cite{lin2022magic3d}. However, optimization-based methods are plagued by long convergence times.
A recent concurrent work~\cite{richardson2023texture} proposes a non-optimization approach with progressive updates from multiple pre-set viewpoints.
In contrast, our method iteratively updates and refines the synthesized textures from automatically selected viewpoint sequences, which minimizes human efforts with designing different viewpoint orders for various geometries.

\section{Method}

The objective of our work is to texture a 3D mesh using a pretrained text-to-image diffusion model. 
In this section, we begin by laying the foundation of the diffusion model in Sec.~\ref{sec: preliminary} and depth-aware inpainting model in Sec.~\ref{sec: inpainting}. 
We then propose a \textit{generate-then-refine} scheme for progressively synthesizing and updating the 3D textures in a coarse-to-fine fashion.
In the progressive texture generation (Sec.~\ref{sec: generation}), we paint the visible regions of the input geometry in an incremental fashion, following a sequence of predefined viewpoints. To ensure local and global consistency, we incorporate a view partition to guide the depth-aware inpainting objectives. 
Subsequently, we introduce an automatic viewpoint selection mechanism  (Sec.~\ref{sec: refinement}) to perform texture refinement and address any issues of texture stretching and inconsistency.

\subsection{Preliminary}
\label{sec: preliminary}

We use a Denoising Diffusion Probabilistic Model (DDPM)~\cite{ho2020denoising} as the generative model. 
Specifically, to avoid high computational overhead, we adopt the latent diffusion model~\cite{Rombach_2022_CVPR}, where an input image $x_0$ is first encoded into latent code $z_0$ before the diffusion process. 
The forward pass follows a Markov Chain to gradually add noise to the input latent code $z_0$ towards the white Gaussian noise $\mathcal{N}(0, 1)$. At each step in the forward pass, the noised latent code $z_t$ is obtained by adding a noise variance $\beta_t$ to the previous latent code $z_{t-1}$ scaled with $\sqrt{1-\beta_t}$:

\begin{equation}
    z_t \sim \mathcal{N}(\sqrt{1-\beta_t} z_{t-1}, \beta_t\textbf{I}).
\end{equation}
The independence property enables direct transformation of the noised latent code $z_t$ at an arbitrary time step $t$ from the input latent $z_0$ via:
\begin{equation}
    z_t \sim \mathcal{N}(\sqrt{\bar{a}}_t z_0, (1-\bar{a}_t)\textbf{I}),
\end{equation}
where $\bar{a}_t$ is the total noise variance, which can be calculated by~$\sum_{t=i}^{T}(1-\beta_{t})$ from the noise $\beta_t$ added to the input latent code $z_0$ at each time step.

During inference, the latent estimation $\hat{z}_{t-1}$ for the next time step $t-1$ is obtained by predicting $\mu_\theta(z_t, t)$ and $\sigma_\theta(z_t, t)$ of a Gaussian distribution:

\begin{equation}
    \hat{z}_{t-1} \sim \mathcal{N}(\mu_\theta(z_t, t), \sigma_\theta(z_t, t))
\end{equation}

\paragraph{Denoising strength.}

To prevent complete randomness during the diffusion process, we introduce a scaling factor $\gamma, 0 < \gamma \le 1$, which controls the number of diffusion steps. We assume that a white Gaussian noise $\mathcal{N}(0, 1)$ can be obtained by adding noise to the input latent code $z_0$ through $T$ steps, and the final denoised latent estimation $\hat{z}_{0}$ is fully governed by the pure noise. By applying the scaling factor, we can start denoising the latent code at time step $\gamma T$ to guide the final latent code with the original image information. This technique is applied to refine the previously generated image contents.

\subsection{Depth-Aware Image Inpainting}
\label{sec: inpainting}

The core of the texture synthesis lies in painting the missing regions on the mesh surface. 
The generated texture is expected to be highly faithful to the mesh geometry and the input text. 
To achieve this, we build our method on a pre-trained depth-to-image model~\cite{stablediffusion, zhang2023adding} that can produce high-quality images from text while being consistent with depth cues. 
However, since the Depth2Image model is designed to generate entire images, we need to use an inpainting mask to guide the sampling process. 
This mask provides explicit hints of which regions to generate or keep fixed, similar to the denoising guidance strategy in RePaint~\cite{lugmayr2022repaint}. 

To condition the denoising on the known regions of the input, we inject a generation mask $\mathcal{M}$ into the sampling steps. This mask explicitly blends the noised latent code $z_t$ and the denoised latent estimation $\hat{z}_t$ as follows:

\begin{equation}
    \hat{z}_t = \hat{z}_t \odot \mathcal{M} + z_t \odot (1 - \mathcal{M}).
\end{equation}
We then decode the final denoised latent estimation $\hat{z}_0$ to the final output image $\mathcal{O}$.

\begin{figure}[!t]
    \centering
    \includegraphics[width=0.99\linewidth]{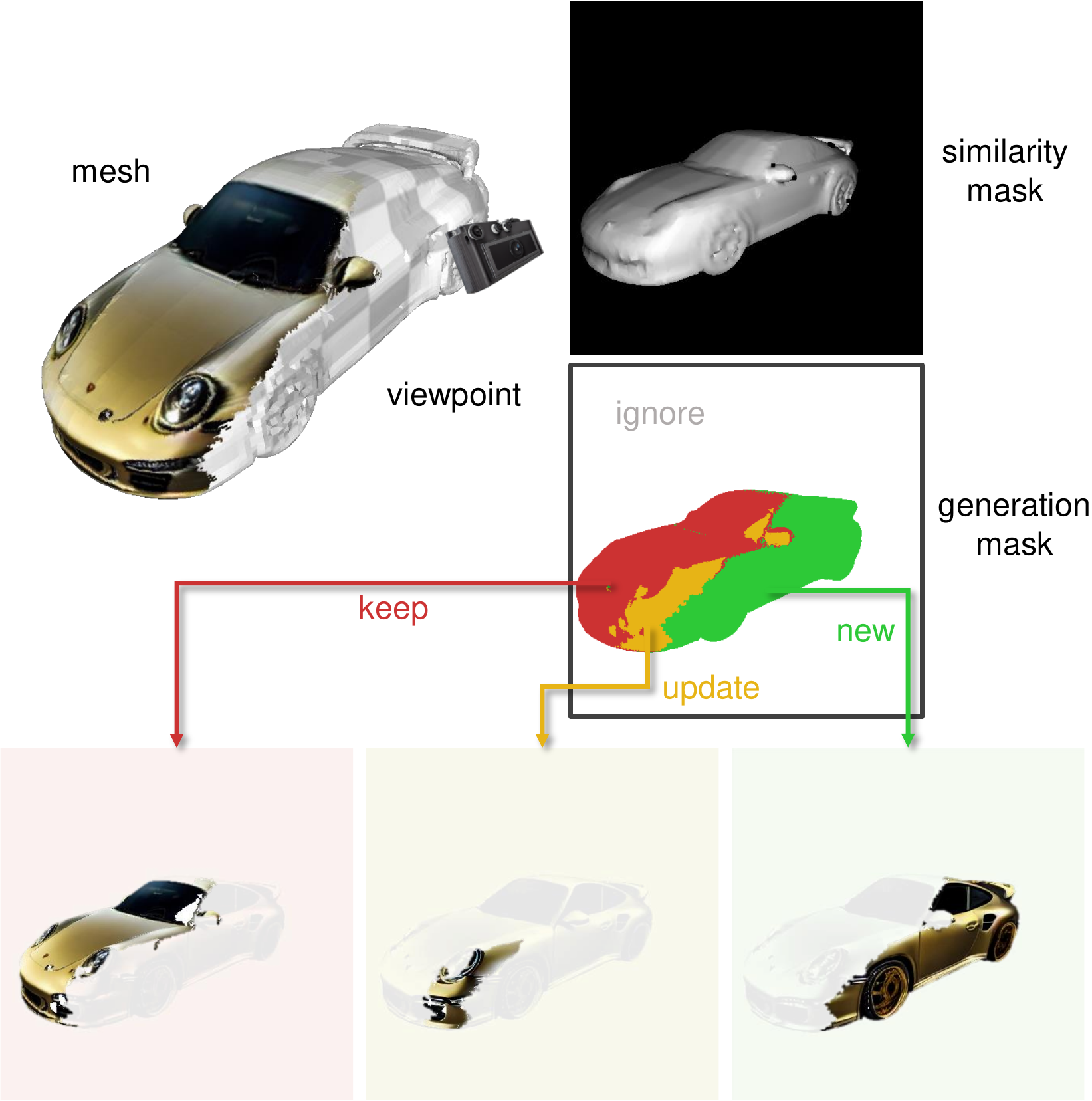}
    \caption{
    We dynamically partition the current view into a generation mask to guide the depth-aware inpainting model. For the ``new'' region, we denoise the new object appearance from white Gaussian noise. For the ``update'' region, we refine the previous texture by denoising the partially noised image segments. We freeze the texture in the ``keep'' region for this view.}
    \label{fig:generation}
\end{figure}
\subsection{Progressive Texture Generation}
\label{sec: generation}

With the customized Depth2Image model,  we are able to paint the object with a high quality image from a single view.
To synthesize the appearance of the input geometry, we project the generated 2D views onto the texture space of a normalized 3D object with proper UV parameterization. 
Assuming Y-axis as the up-axis in the world coordinate system, we define the viewpoint as $v=(\theta, \phi, r)$, where $\theta$ is the azimuth angle with respect to the Z-axis, $\phi$ is the viewpoint elevation angle with respect to the XZ-plane, and $r$ is the distance between the viewpoint and the origin.

As shown in Fig.~\ref{fig:progressive}, we start by generating the visible but missing texture in an initial viewpoint $v^0$. We render the object to a depth map $\mathcal{D}^0$ and a generation mask $\mathcal{M}^0$, and then use a customized Depth2Image diffusion model with $\mathcal{D}^0$ as input and $\mathcal{M}^0$ as extra guidance to generate a colored image $\mathcal{O}^0$. 
We then back-project the image $\mathcal{O}^0$ to the visible part of the texture $\mathcal{T}^0$. 
In the subsequent steps, we progressively diffuse the colored images $\mathcal{O}^k$ and back-project them to the texture $\mathcal{T}^k$ through a sequence of viewpoints.

We notice that directly inpainting the missing regions on mesh surface often results in inconsistency issues. 
The issue is mainly caused by the stretched artifacts that occur when the 2D views are projected back onto the curved surface of the mesh.
Therefore, we design a dynamic view partitioning strategy to guide the inpainting process with respective generation objectives $\mathcal{M}$ and denoising strengths $\gamma$.

\paragraph{Dynamic view partitioning}

For all viewpoints $\mathcal{V}=v_i, i=1, ..., N$, we render the similarity mask $\mathcal{S}^i$ for each viewpoint $v^k$ and map those values to the texture space. 
Each pixel in a similarity mask represents the reversed normalized value of the cosine similarity between the normal vectors of the visible faces and the view direction (ranging from 0 to 1). A pixel with a value of $1$ indicates that the corresponding face is perpendicular to the view direction. For simplicity, we set the background to $0$ . In summary, these masks indicate the extent to which a face is rotated away from the viewpoint.

Based on the similarity mask $\mathcal{S}^k$ at step $k$, we segment the rendered view into a \textbf{generation mask} $\mathcal{M}^k$, including the following $4$ regions, as shown in Fig.~\ref{fig:generation}: 
1) \textbf{New}: This region contains pixels that have not yet been textured. We inpaint this region from pure white Gaussian noise, i.e. with denoising strength $1$.
2) \textbf{Update}: This region contains pixels that have been textured, but the corresponding similarity score in $\mathcal{S}^k$ is greater than all other views. This indicates that those pixels are being observed in a better angle. Therefore, we update this region with a moderate denoising strength $\gamma_g$ to avoid stretched appearance.
3) \textbf{Keep}: Pixels in this region have been textured, but the corresponding similarity score in $\mathcal{S}^k$ is not the highest among all other views.These pixels have already been observed from a better angle, so we keep them fixed.
4) \textbf{Ignore}: This region contains pixels that belong to the background and are irrelevant to the process, so we ignore them throughout the entire process.

While the generation mask helps guide the texture inpainting process with accurate generation objectives and appropriate denoising strengths, blurriness and stretches can still exist on the mesh surface. This is because the generation mask is limited to a predefined set of viewpoints, and the seams and stretches on the texture are still visible from a novel viewpoint. To address this issue, we propose a texture refinement technique with an automatic viewpoint selection strategy, which is described in the next section.

\begin{figure}[!t]
    \centering
    \includegraphics[width=0.99\linewidth]{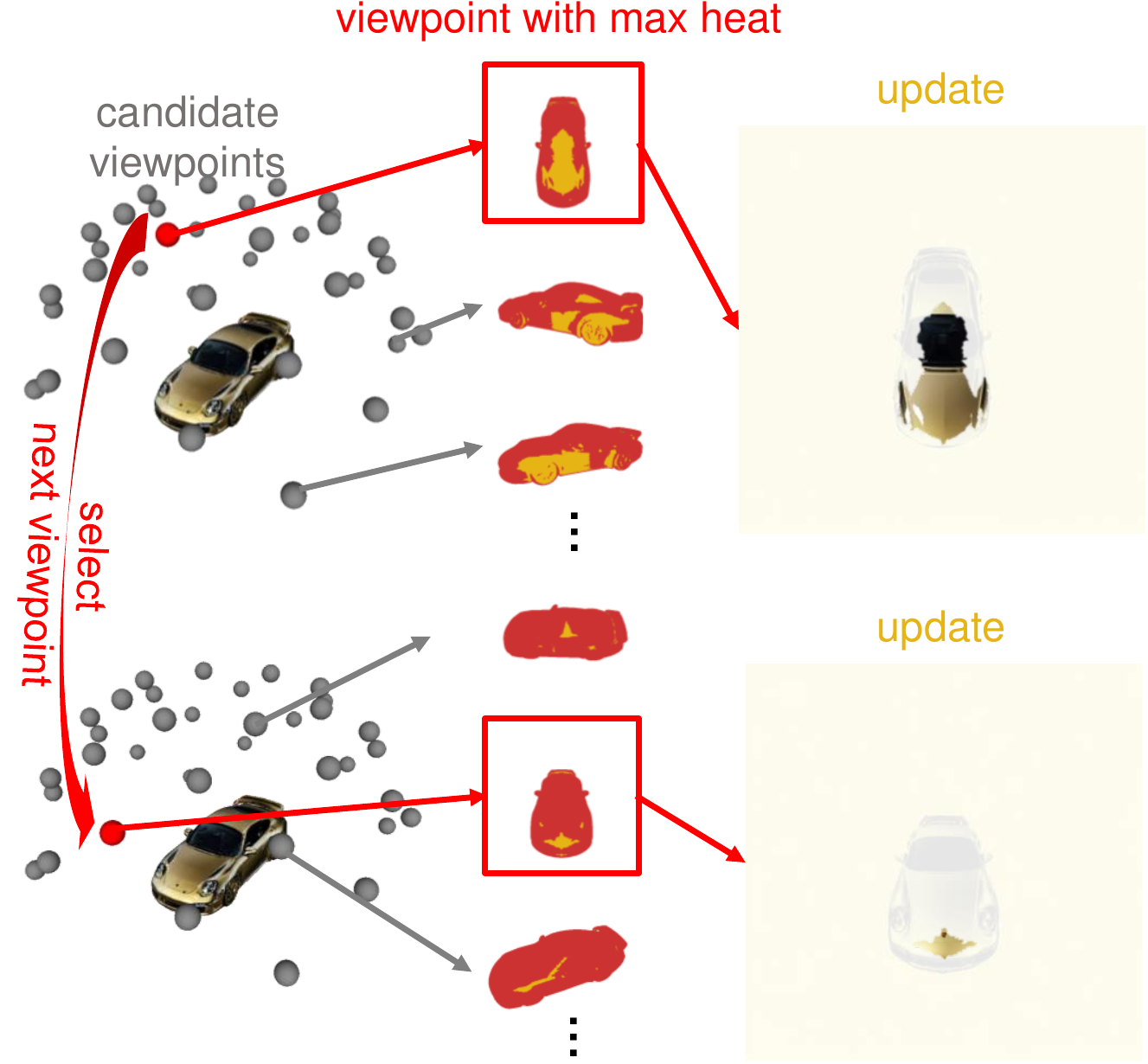}
    \caption{
    In the refinement stage, the sequence of viewpoints are automatically determined by selecting the viewpoint with the maximum normalized area of the ``update'' region at each step. We update 2D views in the ``update'' regions with a moderate diffusion denoising strength. The updated object appearance is then back-projected to the texture space at the end of each refinement step.}
    \label{fig:selection}
\end{figure}

\begin{figure*}[!h]
    \centering
    \includegraphics[width=0.99\linewidth]{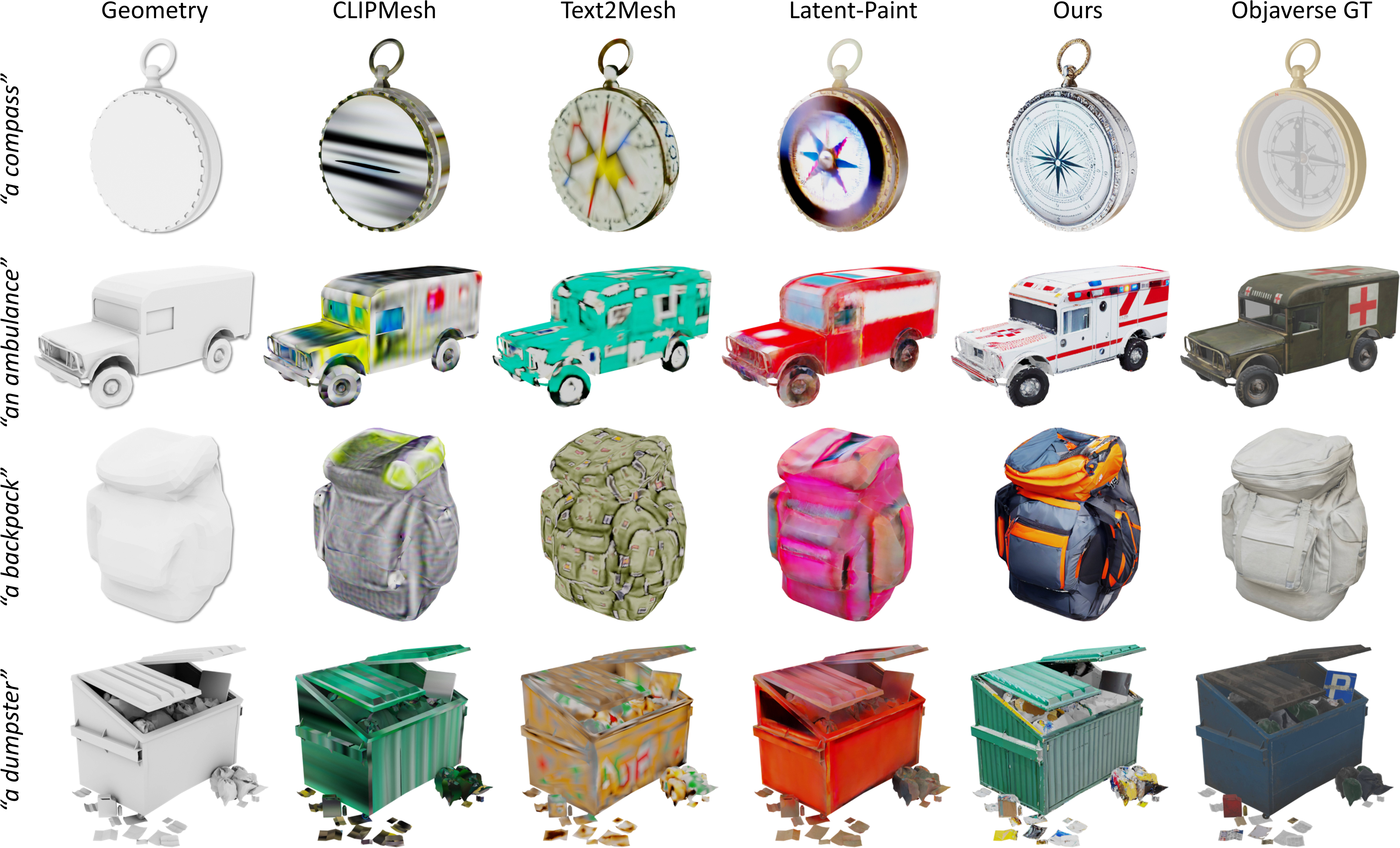}
    \caption{
    Qualitative comparisons on Objaverse.
    We compare our textured mesh against CLIPMesh~\cite{mohammad2022clip}, Text2Mesh~\cite{michel2022text2mesh}, Latent-Paint~\cite{metzer2022latent}, and the original textures from Objaverse. In comparison with the baselines, our method produces more consistent and detailed 3D textures with respect to the input geometries. Image best viewed in color.
    }
    \label{fig:qualitative}
\end{figure*}

\subsection{Texture Refinement with Automatic Viewpoint Selection}
\label{sec: refinement}

To remove the synthesis artifacts, a straightforward solution is to increase the number of viewpoints. 
However, the optimal viewpoint sequence can vary for different object geometries, making it difficult to manually pre-set the viewpoint sequence for massive synthesis targets.
To address these challenges, we propose an automatic viewpoint selection strategy that effectively prevents stretches and seams, as illustrated in Fig.~\ref{fig:selection}. 
We densely define a set of refinement viewpoints $\mathcal{V}=v_i, i=1, ...,K$, where $K$ is larger than $N$. 
To distribute the refinement viewpoints evenly, we scatter them on a hemisphere, taking into account that objects are rarely observed from the bottom-up view.

Assuming that an initial texture has been applied to the object, the refinement process begins by segmenting the generation masks $\mathcal{M}$ using the similarity masks $\mathcal{S}$ from all available viewpoints. For each of the $K$ refinement viewpoints in $\mathcal{V}$, we calculate a view heat $h^i$ from the corresponding generation mask $\mathcal{M}_i$, which represents the normalized area of the ``update'' region with respect to the current visible area of the object. The viewpoint $v_i$ that maximizes the view heat is then selected by
$\operatorname*{arg\,max}_i h_i = \frac{1}{N_p}\sum_{k=1}^{N_p}w_k$
where $N_p$ is the total number of the non-background pixels, and $w_k$ is the scaling factor for the segments in the generation mask. In order to let views with relatively more areas for updating, $w_k$ for the ``update'' region is set to be bigger than that for the ``keep'' region.
We dynamically select the next view with the highest view heat for updating. To avoid conflicts with the previously generated textures, we update the ``update'' regions with a mild denoising strength $\gamma_r$, which preserves the original appearance cues.

\section{Results}

\subsection{Implementation Details}

We apply the Depth2Image model from Stable Diffusion v2~\cite{stablediffusion} as our generation backbone. The denoising strength $\gamma_g$ and $\gamma_r$ are set as $0.5$ and $0.3$ for the generation and refinement stages, respectively. We define $6$ axis-aligned principles viewpoints for generation, and in total $36$ viewpoints for refinement, among which we dynamically select only $20$ views to reduce time cost. Each synthesis process takes around 15 minutes to complete on an NVIDIA RTX A6000. Our implementation uses the PyTorch~\cite{paszke2017automatic} framework with PyTorch3D~\cite{ravi2020pytorch3d} used for rendering and texture projection.

\subsection{Experiment Setup}

\paragraph{Data.}

We evaluate our method on a subset of textured meshes from the Objaverse~\cite{objaverse} dataset. We first sample 3 random meshes from each category. To ensure the quality of the input meshes, we manually filter out thin or unrecognizable meshes, such as ``strainer'', ``sweatpants'', and ''legging'', meshes with too simplistic textures, and meshes that do not correspond with their assigned categories. For the purpose of reducing processing time, we also remove over-triangulated and scanned objects. After this curation, there are in total $410$ high quality textured meshes across $225$ categories for the experiments. Note that the original textures are only used for the evaluation.  
To compare with GAN-based category-specific approaches, we also report results on the ``car'' objects from the ShapeNet dataset~\cite{chang2015shapenet}. In particular, we use the $300$ meshes from the test set used in~\cite{siddiqui2022texturify}.

\paragraph{Baselines.}

We compare our method against the following state-of-the-art text-driven texture synthesis method: 1) \textbf{Text2Mesh}~\cite{michel2022text2mesh}, a neural pipeline that directly optimizes the textures and geometries via a CLIP-based optimization objective. We remove the displacement prediction so that only the surface RGBs are optimized. 2) \textbf{CLIPMesh}~\cite{mohammad2022clip}, a CLIP-based pipeline that deforms a sphere and optimizes the surface RGBs. Similar to Text2Mesh, we remove the shape deformation branch, and the texture colors are directly optimized on the surface of a given geometry. 3) \textbf{Latent-Paint}~\cite{metzer2022latent}, a texture generation variant of the NeRF-based 3D object generation pipeline Latent-NeRF~\cite{metzer2022latent}. It explicitly operates on a given texture map using Stable Diffusion as a prior. 
In addition to text-guided methods, we also compare with category-specific GAN-based approaches, including Texture Fields~\cite{oechsle2019texture}, SPSG~\cite{dai2021spsg}, LTG~\cite{yu2021learning}, and Texturify~\cite{siddiqui2022texturify}.

\paragraph{Evaluation metrics.}

We evaluate the generated textures via commonly used image quality and diversity metrics for generative models. Specifically, we report the Frechet Inception Distance (FID)~\cite{heusel2017gans} and Kernel Inception Distance (KID $\times 10^{-3}$ )~\cite{binkowski2018demystifying}. The generated image distribution for these metrics consists of renders of each mesh with the synthesized textures from 20 fixed viewpoints at a resolution of $512 \times 512$. For experiments on Objaverse dataset, the real distribution comprises renders of the meshes with the same settings using their artist designed textures. For experiments on ShapeNet cars, we use the $18991$ background segmented images from CompCars dataset~\cite{yang2015large}.

\subsection{Quantitative results}

In Tab.~\ref{tab:quantitatives}, we compare our method against the previous SOTA text-driven texture synthesis methods on Objaverse objects. As input, we uniformly feed template texts ``\textit{a \textlangle category\textrangle}'' to the models. Quantitatively, our method outperforms all baselines by a significant margin ($19\%$ improvement in FID and $26\%$ improvement in KID). Such improvements demonstrate that our method is more capable of generating more realistic textures on various object geometries from numerous categories. 
To demonstrate the effectiveness of our method against the GAN-based approaches on category-specific data, we report experiment results on ShapeNet ``car'' category in Tab.~\ref{tab:texturify}. Notably, our method achieves superior performance over the previous GAN-based SOTA texture synthesis method Texturify, improving by $21\%$ in FID and $12\%$ in KID. This indicates our method is more effective with synthesizing realistic textures than GAN based approaches that were trained on specific categories.

\begin{figure}
    \centering
    \includegraphics[width=\linewidth]{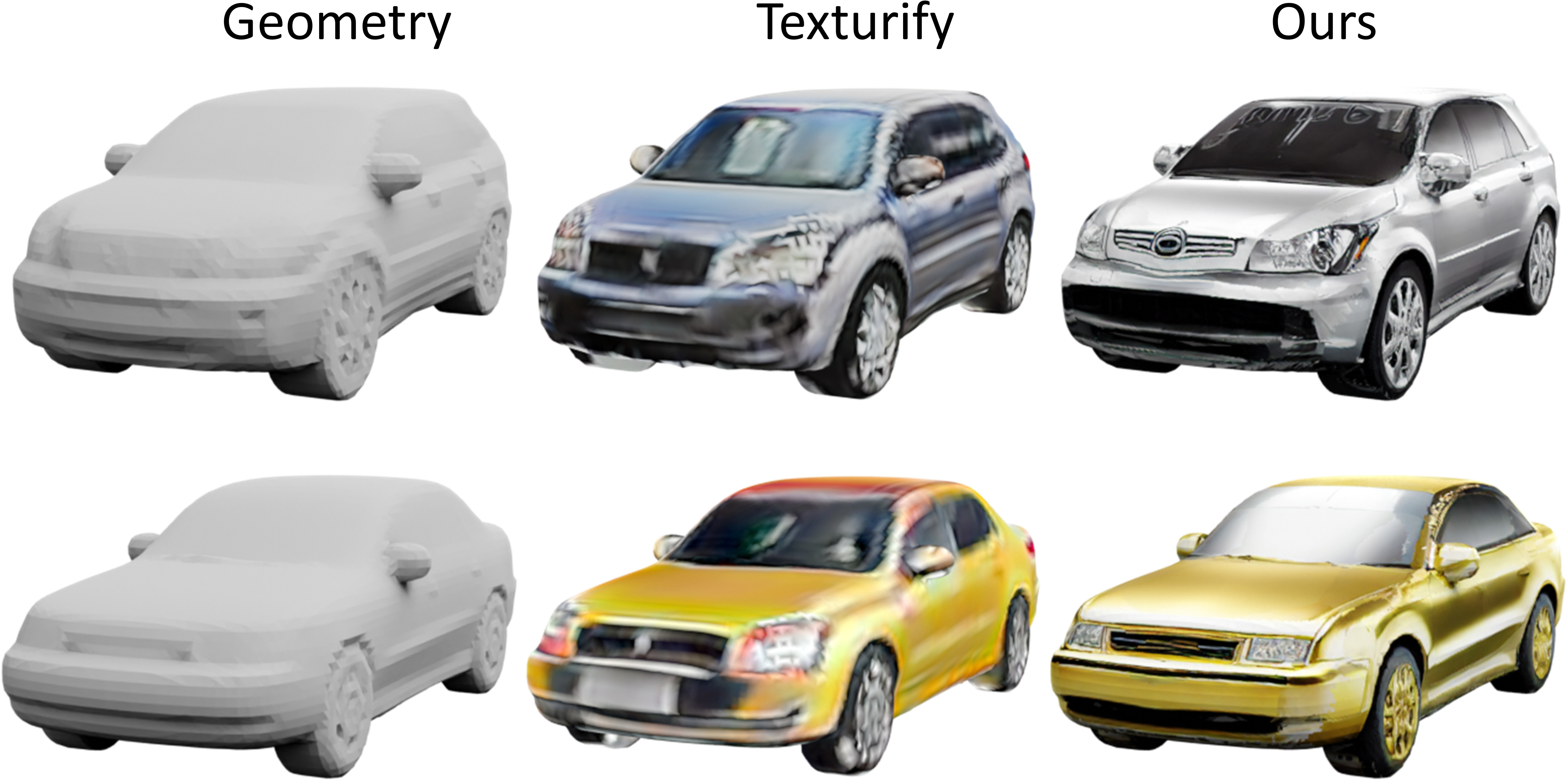}
    \caption{
    Qualitative comparisons on ShapeNet car.
    Our method generates sharper and more coherent textures with respect to the geometries compared to the state-of-the-art GAN-based method. 
    }

    \label{fig:cars}
\end{figure}

\paragraph{User study.}

We conduct user study to analyze the quality of the synthesized textures and their fidelity to the input text prompts. 
For each baseline method, we randomly show the users 5 pair of renders from the baseline and our method. The users are requested to choose the one that is more realistic and closer to the text prompts.
More details of the questionnaire can be found in the supplemental. In the end, we receive 604 responses across 41 users. The collected preferences are reported in Tab.~\ref{tab:user}. In comparison to CLIPMesh and Text2Mesh, our method is clearly preferred by the users with preference rate $83.92\%$ and $76.47\%$, respectively. Besides that, more users ($64.18\%$) lean towards our method over the competitive baseline Latent-Paint. As can be seen, our method demonstrates the effectiveness in generating high quality textures that are favored by human users.

\begin{table}[!t]
    \centering
        \begin{tabular}{l|ll}
            \toprule
            Method & FID $\downarrow$ & KID (x$10^{-3}$) $\downarrow$ \\
            \midrule
            Text2Mesh~\cite{michel2022text2mesh} & 45.38 $_{\;\text{\textcolor{gray}{(+9.7)}}}$ & 10.40 $_{\;\text{\textcolor{gray}{(+2.7)}}}$ \\
            CLIPMesh~\cite{mohammad2022clip} & 43.25 $_{\;\text{\textcolor{gray}{(+7.6)}}}$ & 12.52 $_{\;\text{\textcolor{gray}{(+4.8)}}}$ \\
            Latent-Paint~\cite{metzer2022latent} & 43.87 $_{\;\text{\textcolor{gray}{(+8.1)}}}$ & 11.43 $_{\;\text{\textcolor{gray}{(+3.7)}}}$ \\
            \midrule
            \ARCH (Ours) & \textbf{35.68} & \textbf{7.74} \\
            \bottomrule
        \end{tabular}
    \caption{
    Quantitative comparisons on Objaverse subset.
    Our method performs favaorably against state-of-the-art text-driven texture synthesis methods.
    }
    \label{tab:quantitatives}
\end{table}

\begin{table}[!t]
    \centering
        \begin{tabular}{l|ll}
            \toprule
            Method & FID $\downarrow$ & KID (x$10^{-3}$) $\downarrow$ \\
            \midrule
            Texture Fields~\cite{oechsle2019texture} & 177.15 $_{\;\text{\textcolor{gray}{(+130.2)}}}$ & 17.14 $_{\;\text{\textcolor{gray}{(+12.8)}}}$ \\
            SPSG~\cite{dai2021spsg} & 110.65 $_{\;\text{\textcolor{gray}{(+63.7)}}}$ & 9.59 $_{\;\;\;\text{\textcolor{gray}{(+5.2)}}}$ \\
            LTG~\cite{yu2021learning} & 70.76 $_{\;\;\;\text{\textcolor{gray}{(+23.8)}}}$ & 5.72 $_{\;\;\;\text{\textcolor{gray}{(+1.4)}}}$ \\
            Texturify~\cite{siddiqui2022texturify} & 59.55 $_{\;\;\;\text{\textcolor{gray}{(+12.6)}}}$ & 4.97 $_{\;\;\;\text{\textcolor{gray}{(+0.6)}}}$ \\
            \midrule
            \ARCH (Ours) & \textbf{46.91} & \textbf{4.35} \\
            \bottomrule
        \end{tabular}
    \caption{
    Quantitative comparison on the ShapeNet cars. Our method outperform state-of-the-art category-specific GAN-based methods  by a significant margin.
    }
    \label{tab:texturify}
\end{table}

\begin{table}[!t]
    \centering
        \begin{tabular}{c|ccc}
            \toprule
             & CLIPMesh $\uparrow$ & Text2Mesh $\uparrow$ & Latent-Paint $\uparrow$ \\
            \midrule
            Ours & 83.92\% & 76.47\% & 64.18\% \\
            \bottomrule
        \end{tabular}
    \caption{Percentage of users who prefer our method over the baselines in a user study with 604 responses across 41 participants. Our method is shown to be more favored by human users.
    }
    \label{tab:user}
\end{table}

\subsection{Qualitative analysis}

We compare our qualitative results on Objaverse objects against text-driven baselines in Fig.~\ref{fig:qualitative}. In comparison with CLIP-based methods CLIPMesh and Text2Mesh, our method generates more realistic and consistent textures. 
In particular, 
CLIPMesh generates sketchy textures while Text2Mesh produces repetitive patterns. 
Latent-Paint outputs a consistent texture capturing the semantics of the object well, but the results are often quite blurry.
Clearly, our method can synthesize more consistent textures with cleaner and richer local details.
We also compare our textures with GAN-based generation approach on category-specific objectives. In Fig.~\ref{fig:cars}, we show the textures for ShapeNet cars of our methods and Texturify. Notably, our textures have a much cleaner appearance and provide more details with respect to the input geometries.

\subsection{Ablation studies}

\begin{figure}[!t]
    \centering
    \includegraphics[width=0.99\linewidth]{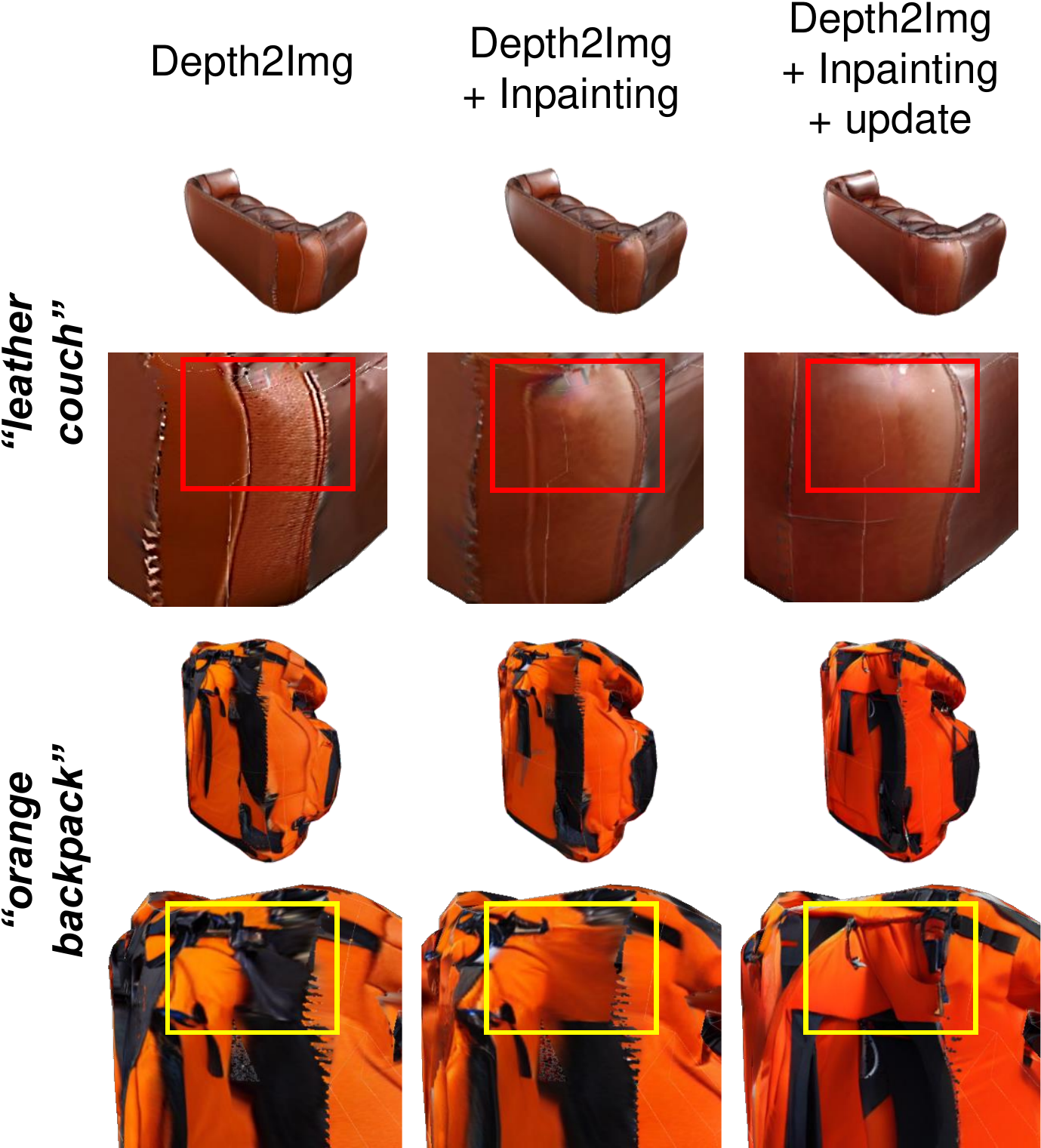}
    \caption{
    The proposed inpainting and update technique can effectively generate more consistent textures and eliminate stretched and blurry artifacts. Note that there is no refinement in this ablation.
    }
    \label{fig:inpainting}
\end{figure}

\paragraph{Does depth-aware inpainting and updating help?}

\begin{table}[!t]
    \centering
    \resizebox{\linewidth}{!}{
        \begin{tabular}{ccc|cc}
            \toprule
            w/ Depth2Img & w/ inpainting & w/ update & FID $\downarrow$ & KID (x$10^{-3}$) $\downarrow$ \\
            \midrule
            \checkmark & \text{\sffamily x} & \text{\sffamily x} & 39.88 & 9.78 \\
            \checkmark & \checkmark & \text{\sffamily x} & 38.19 & 9.11 \\
            \checkmark & \checkmark & \checkmark & \textbf{37.09} & \textbf{8.78} \\
            \bottomrule
        \end{tabular}
    }
    \caption{
    Effect of components in the generation stage.
    We quantitatively ablate the efficacy of each component. Applying the inpainting and update scheme effectively improves the quality of the synthesized textures.
    }
    \label{tab:inpainting}
\end{table}

We show in Fig.~\ref{fig:inpainting} that the depth-aware inpainting is essential for producing high quality textures. 
In particular, the plain Depth2Img model often struggles to produce consistent appearances due to the governance of random noise. When the inpainting scheme is applied, the produced textures are more consistent. However, the textures still appear to be stretched and blurry over the curved mesh surface. These artifacts are amended by the texture updating scheme from the better viewing angles. The effectiveness of the depth-aware inpainting and the updating scheme is reflected in the improved FID and KID scores in Tab.~\ref{tab:inpainting}.

\paragraph{Does viewpoint selection in refinement stage help?}

\begin{figure}[!t]
    \centering
    \includegraphics[width=0.99\linewidth]{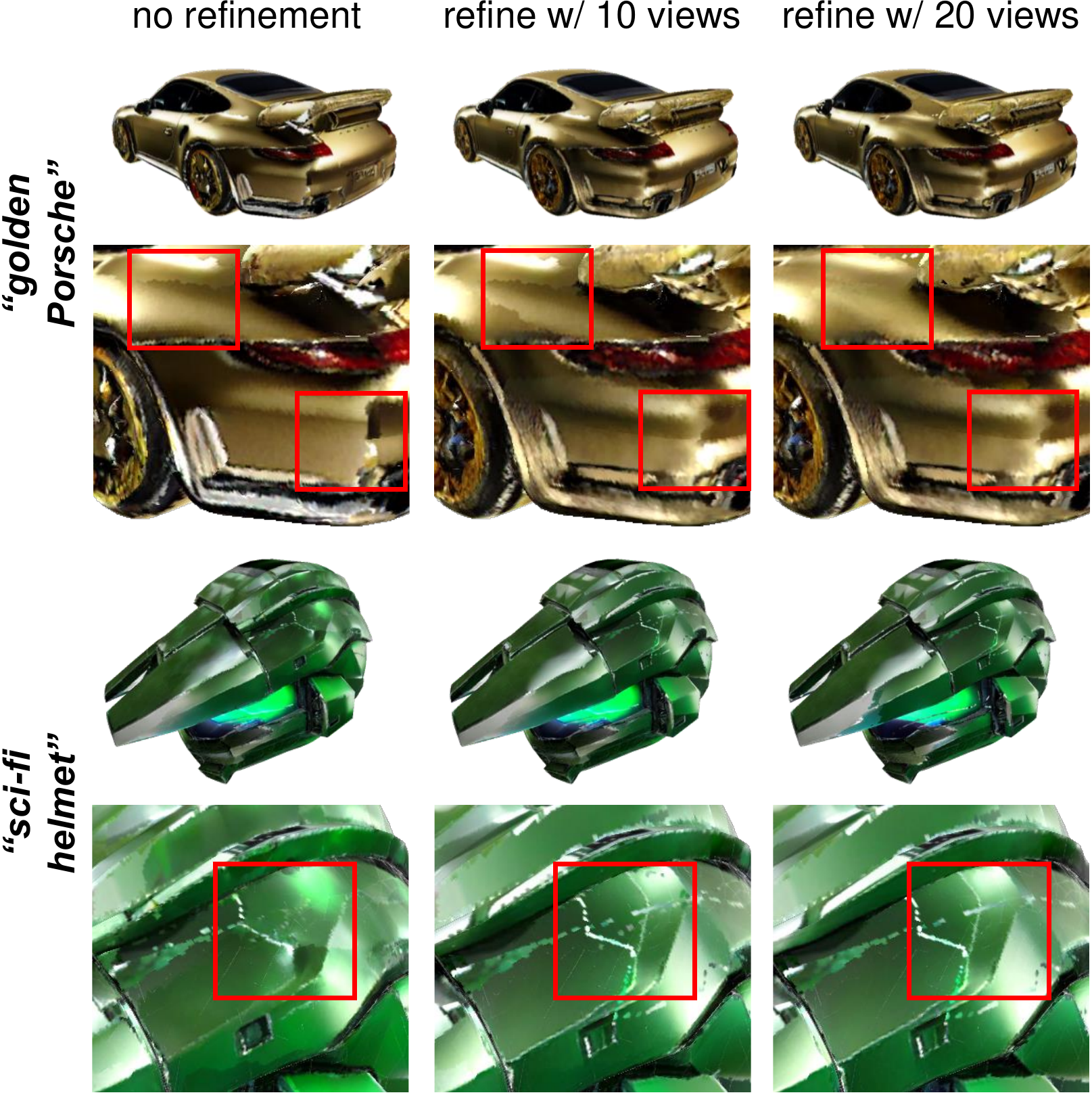}
    \caption{
    The proposed automatic viewpoint selection method further improves the texture quality by gradually removing the remaining artifacts from the generation stage.
    }
    \label{fig:refinement}
\end{figure}

We compare our results with different refinement settings in Tab.~\ref{tab:refinement}. When the input geometries are painted with initial textures from the generation stage, the blurry artifacts and projection seams are usually not eliminated due to a limited number of viewpoints. As can be seen in Fig.~\ref{fig:refinement}, such flaws can be minimized by refining with more viewpoints. We also showcase the effectiveness of the automatic viewpoint selection technique, as the refinement process does not require any manual efforts with defining and fine-tuning the viewpoint sequence for different shapes.

\begin{table}[!t]
    \centering
        \resizebox{\linewidth}{!}{
            \begin{tabular}{l|ccccc}
                \toprule
                \# views & 0 & 5 & 10 & 15 & 20 \\
                \midrule
                $\downarrow$ FID & 37.09 & 36.67 & 36.39 & 35.98 & \textbf{35.68} \\
                $\downarrow$ KID (x$10^{-3}$) & 8.78 & 8.31 & 8.12 & 7.98 & \textbf{7.74} \\
                \bottomrule
            \end{tabular}
        }
    \caption{
    We quantitatively study the effect of selecting different number of viewpoints in the refinement stage. Refining the synthesis results with more viewpoints improves the texture quality.
    }
    \label{tab:refinement}
\end{table}

\subsection{Limitations.}

While our method has shown the capability to produce high-quality 3D textures, we have observed that it tends to produce textures with shading effects from the diffusion backbone. Although this issue can be addressed by carefully fine-tuning the input prompts, doing so requires additional human engineering effort and may not scale well to massive generation targets. One potential solution is to fine-tune the diffusion model to remove the shading from textures. We acknowledge this challenge and leave it to future work to explore this possibility.
\section{Conclusion}

In this paper, we present a novel method,~\ARCH, for synthesizing high quality textures for 3D meshes from the given text prompts. 
Our approach leverages a depth-aware image inpainting diffusion model to progressively generate high-resolution partial textures from multiple viewpoints.
To avoid accumulating inconsistent and stretched artifacts across viewpoints, we dynamically segment the rendered view into a generation mask, which effectively guides the diffusion model to generate and update the corresponding partial textures. 
Furthermore, we propose an automatic viewpoint sequence generation scheme that utilizes the generation mask to automatically determine the next best view for refining the generated textures.
Extensive experiments demonstrate that our method can effectively synthesize consistent and highly detailed 3D textures for various object geometries without extra manual effort.
Overall, we hope our work can inspire more future research in the area of text-to-3D synthesis.

\section*{Acknowledgements}
This work was supported by the ERC Starting Grant Scan2CAD (804724) and the German Research Foundation (DFG)
Research Unit “Learning and Simulation in Visual Computing”.
We further thank Guy Gafni and Manuel Dahnert for the helpful discussions, and Angela Dai for the video voice-over.

% \clearpage
{\small
\bibliographystyle{ieee_fullname}
\bibliography{reference}

\begin{thebibliography}{10}\itemsep=-1pt

\bibitem{stablediffusion}
Stable diffusion.
\newblock \url{https://github.com/Stability-AI/stablediffusion}, 2022.

\bibitem{abdal20233davatargan}
Rameen Abdal, Hsin-Ying Lee, Peihao Zhu, Menglei Chai, Aliaksandr Siarohin,
  Peter Wonka, and Sergey Tulyakov.
\newblock 3davatargan: Bridging domains for personalized editable avatars.
\newblock In {\em CVPR}, 2023.

\bibitem{achlioptas2018learning}
Panos Achlioptas, Olga Diamanti, Ioannis Mitliagkas, and Leonidas Guibas.
\newblock Learning representations and generative models for 3d point clouds.
\newblock In {\em ICML}, 2018.

\bibitem{binkowski2018demystifying}
Miko{\l}aj Bi{\'n}kowski, Danica~J Sutherland, Michael Arbel, and Arthur
  Gretton.
\newblock Demystifying mmd gans.
\newblock In {\em ICLR}, 2018.

\bibitem{EG3D}
Eric~R. Chan, Connor~Z. Lin, Matthew~A. Chan, Koki Nagano, Boxiao Pan,
  Shalini~De Mello, Orazio Gallo, Leonidas Guibas, Jonathan Tremblay, Sameh
  Khamis, Tero Karras, and Gordon Wetzstein.
\newblock Efficient geometry-aware {3D} generative adversarial networks.
\newblock In {\em CVPR}, 2022.

\bibitem{piGAN}
Eric~R Chan, Marco Monteiro, Petr Kellnhofer, Jiajun Wu, and Gordon Wetzstein.
\newblock pi-gan: Periodic implicit generative adversarial networks for
  3d-aware image synthesis.
\newblock In {\em CVPR}, 2021.

\bibitem{chang2015shapenet}
Angel~X Chang, Thomas Funkhouser, Leonidas Guibas, Pat Hanrahan, Qixing Huang,
  Zimo Li, Silvio Savarese, Manolis Savva, Shuran Song, Hao Su, et~al.
\newblock Shapenet: An information-rich 3d model repository.
\newblock {\em arXiv preprint arXiv:1512.03012}, 2015.

\bibitem{chen2020scanrefer}
Dave~Zhenyu Chen, Angel~X Chang, and Matthias Nie{\ss}ner.
\newblock {ScanRefer}: {3D} object localization in {RGB-D} scans using natural
  language.
\newblock In {\em European Conference on Computer Vision}, pages 202--221.
  Springer, 2020.

\bibitem{chen2022unit3d}
Dave~Zhenyu Chen, Ronghang Hu, Xinlei Chen, Matthias Nie{\ss}ner, and Angel~X
  Chang.
\newblock Unit3d: A unified transformer for 3d dense captioning and visual
  grounding.
\newblock {\em arXiv preprint arXiv:2212.00836}, 2022.

\bibitem{chen2022d3net}
Dave~Zhenyu Chen, Qirui Wu, Matthias Nie{\ss}ner, and Angel~X Chang.
\newblock D 3 net: A unified speaker-listener architecture for 3d dense
  captioning and visual grounding.
\newblock In {\em Computer Vision--ECCV 2022: 17th European Conference, Tel
  Aviv, Israel, October 23--27, 2022, Proceedings, Part XXXII}, pages 487--505.
  Springer, 2022.

\bibitem{chen2021scan2cap}
Zhenyu Chen, Ali Gholami, Matthias Nie{\ss}ner, and Angel~X Chang.
\newblock Scan2cap: Context-aware dense captioning in rgb-d scans.
\newblock In {\em Proceedings of the IEEE/CVF Conference on Computer Vision and
  Pattern Recognition}, pages 3193--3203, 2021.

\bibitem{chen2019learning}
Zhiqin Chen and Hao Zhang.
\newblock Learning implicit fields for generative shape modeling.
\newblock In {\em CVPR}, 2019.

\bibitem{cheng2022sdfusion}
Yen-Chi Cheng, Hsin-Ying Lee, Sergey Tulyakov, Alexander Schwing, and Liangyan
  Gui.
\newblock Sdfusion: Multimodal 3d shape completion, reconstruction, and
  generation.
\newblock In {\em CVPR}, 2023.

\bibitem{cheng2022cross}
Zezhou Cheng, Menglei Chai, Jian Ren, Hsin-Ying Lee, Kyle Olszewski, Zeng
  Huang, Subhransu Maji, and Sergey Tulyakov.
\newblock Cross-modal 3d shape generation and manipulation.
\newblock In {\em ECCV}, 2022.

\bibitem{crowson2022vqgan}
Katherine Crowson, Stella Biderman, Daniel Kornis, Dashiell Stander, Eric
  Hallahan, Louis Castricato, and Edward Raff.
\newblock Vqgan-clip: Open domain image generation and editing with natural
  language guidance.
\newblock In {\em ECCV}, 2022.

\bibitem{dai2021spsg}
Angela Dai, Yawar Siddiqui, Justus Thies, Julien Valentin, and Matthias
  Nie{\ss}ner.
\newblock Spsg: Self-supervised photometric scene generation from rgb-d scans.
\newblock In {\em CVPR}, 2021.

\bibitem{objaverse}
Matt Deitke, Dustin Schwenk, Jordi Salvador, Luca Weihs, Oscar Michel, Eli
  VanderBilt, Ludwig Schmidt, Kiana Ehsani, Aniruddha Kembhavi, and Ali
  Farhadi.
\newblock Objaverse: A universe of annotated 3d objects.
\newblock {\em arXiv preprint arXiv:2212.08051}, 2022.

\bibitem{dhariwal2021diffusion}
Prafulla Dhariwal and Alexander Nichol.
\newblock Diffusion models beat gans on image synthesis.
\newblock {\em NeurIPS}, 2021.

\bibitem{esser2021taming}
Patrick Esser, Robin Rombach, and Bjorn Ommer.
\newblock Taming transformers for high-resolution image synthesis.
\newblock In {\em CVPR}, 2021.

\bibitem{gao2022get3d}
Jun Gao, Tianchang Shen, Zian Wang, Wenzheng Chen, Kangxue Yin, Daiqing Li, Or
  Litany, Zan Gojcic, and Sanja Fidler.
\newblock Get3d: A generative model of high quality 3d textured shapes learned
  from images.
\newblock In {\em Advances In Neural Information Processing Systems}, 2022.

\bibitem{GANs}
Ian Goodfellow, Jean Pouget-Abadie, Mehdi Mirza, Bing Xu, David Warde-Farley,
  Sherjil Ozair, Aaron Courville, and Yoshua Bengio.
\newblock Generative adversarial nets.
\newblock {\em NIPS}, 2014.

\bibitem{StyleNeRF}
Jiatao Gu, Lingjie Liu, Peng Wang, and Christian Theobalt.
\newblock Stylenerf: A style-based 3d aware generator for high-resolution image
  synthesis.
\newblock In {\em ICLR}, 2022.

\bibitem{heusel2017gans}
Martin Heusel, Hubert Ramsauer, Thomas Unterthiner, Bernhard Nessler, and Sepp
  Hochreiter.
\newblock Gans trained by a two time-scale update rule converge to a local nash
  equilibrium.
\newblock In {\em NeurIPS}, 2017.

\bibitem{ho2020denoising}
Jonathan Ho, Ajay Jain, and Pieter Abbeel.
\newblock Denoising diffusion probabilistic models.
\newblock {\em NeurIPS}, 2020.

\bibitem{ho2021cascaded}
Jonathan Ho, Chitwan Saharia, William Chan, David~J Fleet, Mohammad Norouzi,
  and Tim Salimans.
\newblock Cascaded diffusion models for high fidelity image generation.
\newblock {\em arXiv preprint arXiv:2106.15282}, 2021.

\bibitem{hu2021unit}
Ronghang Hu and Amanpreet Singh.
\newblock Unit: Multimodal multitask learning with a unified transformer.
\newblock In {\em Proceedings of the IEEE/CVF International Conference on
  Computer Vision}, pages 1439--1449, 2021.

\bibitem{karras2020analyzing}
Tero Karras, Samuli Laine, Miika Aittala, Janne Hellsten, Jaakko Lehtinen, and
  Timo Aila.
\newblock Analyzing and improving the image quality of stylegan.
\newblock In {\em CVPR}, 2020.

\bibitem{kawar2022imagic}
Bahjat Kawar, Shiran Zada, Oran Lang, Omer Tov, Huiwen Chang, Tali Dekel, Inbar
  Mosseri, and Michal Irani.
\newblock Imagic: Text-based real image editing with diffusion models.
\newblock {\em arXiv preprint arXiv:2210.09276}, 2022.

\bibitem{lin2022magic3d}
Chen-Hsuan Lin, Jun Gao, Luming Tang, Towaki Takikawa, Xiaohui Zeng, Xun Huang,
  Karsten Kreis, Sanja Fidler, Ming-Yu Liu, and Tsung-Yi Lin.
\newblock Magic3d: High-resolution text-to-3d content creation.
\newblock {\em arXiv preprint arXiv:2211.10440}, 2022.

\bibitem{lin2023infinicity}
Chieh~Hubert Lin, Hsin-Ying Lee, Willi Menapace, Menglei Chai, Aliaksandr
  Siarohin, Ming-Hsuan Yang, and Sergey Tulyakov.
\newblock Infinicity: Infinite-scale city synthesis.
\newblock {\em arXiv preprint arXiv:2301.09637}, 2023.

\bibitem{lugmayr2022repaint}
Andreas Lugmayr, Martin Danelljan, Andres Romero, Fisher Yu, Radu Timofte, and
  Luc Van~Gool.
\newblock Repaint: Inpainting using denoising diffusion probabilistic models.
\newblock In {\em CVPR}, 2022.

\bibitem{luo2021diffusion}
Shitong Luo and Wei Hu.
\newblock Diffusion probabilistic models for 3d point cloud generation.
\newblock In {\em CVPR}, 2021.

\bibitem{metzer2022latent}
Gal Metzer, Elad Richardson, Or Patashnik, Raja Giryes, and Daniel Cohen-Or.
\newblock Latent-nerf for shape-guided generation of 3d shapes and textures.
\newblock In {\em CVPR}, 2023.

\bibitem{michel2022text2mesh}
Oscar Michel, Roi Bar-On, Richard Liu, Sagie Benaim, and Rana Hanocka.
\newblock Text2mesh: Text-driven neural stylization for meshes.
\newblock In {\em CVPR}, 2022.

\bibitem{mildenhall2020nerf}
Ben Mildenhall, Pratul~P. Srinivasan, Matthew Tancik, Jonathan~T. Barron, Ravi
  Ramamoorthi, and Ren Ng.
\newblock Nerf: Representing scenes as neural radiance fields for view
  synthesis.
\newblock In {\em ECCV}, 2020.

\bibitem{autosdf2022}
Paritosh Mittal, Yen-Chi Cheng, Maneesh Singh, and Shubham Tulsiani.
\newblock {AutoSDF}: Shape priors for 3d completion, reconstruction and
  generation.
\newblock In {\em CVPR}, 2022.

\bibitem{mohammad2022clip}
Nasir Mohammad~Khalid, Tianhao Xie, Eugene Belilovsky, and Tiberiu Popa.
\newblock Clip-mesh: Generating textured meshes from text using pretrained
  image-text models.
\newblock In {\em SIGGRAPH Asia}, 2022.

\bibitem{muller2022diffrf}
Norman M{\"u}ller, Yawar Siddiqui, Lorenzo Porzi, Samuel~Rota Bul{\`o}, Peter
  Kontschieder, and Matthias Nie{\ss}ner.
\newblock Diffrf: Rendering-guided 3d radiance field diffusion.
\newblock {\em arXiv preprint arXiv:2212.01206}, 2022.

\bibitem{nash2020polygen}
Charlie Nash, Yaroslav Ganin, SM~Ali Eslami, and Peter Battaglia.
\newblock Polygen: An autoregressive generative model of 3d meshes.
\newblock In {\em International conference on machine learning}, pages
  7220--7229. PMLR, 2020.

\bibitem{nichol2021improved}
Alexander~Quinn Nichol and Prafulla Dhariwal.
\newblock Improved denoising diffusion probabilistic models.
\newblock In {\em ICML}, 2021.

\bibitem{Giraffe}
Michael Niemeyer and Andreas Geiger.
\newblock Giraffe: Representing scenes as compositional generative neural
  feature fields.
\newblock In {\em CVPR}, 2021.

\bibitem{oechsle2019texture}
Michael Oechsle, Lars Mescheder, Michael Niemeyer, Thilo Strauss, and Andreas
  Geiger.
\newblock Texture fields: Learning texture representations in function space.
\newblock In {\em ICCV}, 2019.

\bibitem{or2022stylesdf}
Roy Or-El, Xuan Luo, Mengyi Shan, Eli Shechtman, Jeong~Joon Park, and Ira
  Kemelmacher-Shlizerman.
\newblock Stylesdf: High-resolution 3d-consistent image and geometry
  generation.
\newblock In {\em CVPR}, 2022.

\bibitem{paszke2017automatic}
Adam Paszke, Sam Gross, Soumith Chintala, Gregory Chanan, Edward Yang, Zachary
  DeVito, Zeming Lin, Alban Desmaison, Luca Antiga, and Adam Lerer.
\newblock Automatic differentiation in pytorch.
\newblock 2017.

\bibitem{patashnik2021styleclip}
Or Patashnik, Zongze Wu, Eli Shechtman, Daniel Cohen-Or, and Dani Lischinski.
\newblock Styleclip: Text-driven manipulation of stylegan imagery.
\newblock In {\em ICCV}, 2021.

\bibitem{poole2022dreamfusion}
Ben Poole, Ajay Jain, Jonathan~T Barron, and Ben Mildenhall.
\newblock Dreamfusion: Text-to-3d using 2d diffusion.
\newblock In {\em ICLR}, 2023.

\bibitem{radford2021learning}
Alec Radford, Jong~Wook Kim, Chris Hallacy, Aditya Ramesh, Gabriel Goh,
  Sandhini Agarwal, Girish Sastry, Amanda Askell, Pamela Mishkin, Jack Clark,
  et~al.
\newblock Learning transferable visual models from natural language
  supervision.
\newblock In {\em ICLR}, 2021.

\bibitem{ramesh2022hierarchical}
Aditya Ramesh, Prafulla Dhariwal, Alex Nichol, Casey Chu, and Mark Chen.
\newblock Hierarchical text-conditional image generation with clip latents.
\newblock {\em arXiv preprint arXiv:2204.06125}, 2022.

\bibitem{ramesh2021zero}
Aditya Ramesh, Mikhail Pavlov, Gabriel Goh, Scott Gray, Chelsea Voss, Alec
  Radford, Mark Chen, and Ilya Sutskever.
\newblock Zero-shot text-to-image generation.
\newblock In {\em ICML}, 2021.

\bibitem{ravi2020pytorch3d}
Nikhila Ravi, Jeremy Reizenstein, David Novotny, Taylor Gordon, Wan-Yen Lo,
  Justin Johnson, and Georgia Gkioxari.
\newblock Accelerating 3d deep learning with pytorch3d.
\newblock {\em arXiv:2007.08501}, 2020.

\bibitem{richardson2023texture}
Elad Richardson, Gal Metzer, Yuval Alaluf, Raja Giryes, and Daniel Cohen-Or.
\newblock Texture: Text-guided texturing of 3d shapes.
\newblock {\em arXiv preprint arXiv:2302.01721}, 2023.

\bibitem{Rombach_2022_CVPR}
Robin Rombach, Andreas Blattmann, Dominik Lorenz, Patrick Esser, and Bj\"orn
  Ommer.
\newblock High-resolution image synthesis with latent diffusion models.
\newblock In {\em CVPR}, 2022.

\bibitem{saharia2022image}
Chitwan Saharia, Jonathan Ho, William Chan, Tim Salimans, David~J Fleet, and
  Mohammad Norouzi.
\newblock Image super-resolution via iterative refinement.
\newblock 2022.

\bibitem{GRAF}
Katja Schwarz, Yiyi Liao, Michael Niemeyer, and Andreas Geiger.
\newblock Graf: Generative radiance fields for 3d-aware image synthesis.
\newblock In {\em NeurIPS}, 2020.

\bibitem{siarohin2023unsupervised}
Aliaksandr Siarohin, Willi Menapace, Ivan Skorokhodov, Kyle Olszewski, Jian
  Ren, Hsin-Ying Lee, Menglei Chai, and Sergey Tulyakov.
\newblock Unsupervised volumetric animation.
\newblock In {\em CVPR}, 2023.

\bibitem{siddiqui2022texturify}
Yawar Siddiqui, Justus Thies, Fangchang Ma, Qi Shan, Matthias Nie{\ss}ner, and
  Angela Dai.
\newblock Texturify: Generating textures on 3d shape surfaces.
\newblock In {\em ECCV}, 2022.

\bibitem{singh2022flava}
Amanpreet Singh, Ronghang Hu, Vedanuj Goswami, Guillaume Couairon, Wojciech
  Galuba, Marcus Rohrbach, and Douwe Kiela.
\newblock Flava: A foundational language and vision alignment model.
\newblock In {\em Proceedings of the IEEE/CVF Conference on Computer Vision and
  Pattern Recognition}, pages 15638--15650, 2022.

\bibitem{skorokhodov3d}
Ivan Skorokhodov, Aliaksandr Siarohin, Yinghao Xu, Jian Ren, Hsin-Ying Lee,
  Peter Wonka, and Sergey Tulyakov.
\newblock 3d generation on imagenet.
\newblock In {\em ICLR}, 2023.

\bibitem{smith2017improved}
Edward~J Smith and David Meger.
\newblock Improved adversarial systems for 3d object generation and
  reconstruction.
\newblock In {\em CoRL}, 2017.

\bibitem{wang2022ofa}
Peng Wang, An Yang, Rui Men, Junyang Lin, Shuai Bai, Zhikang Li, Jianxin Ma,
  Chang Zhou, Jingren Zhou, and Hongxia Yang.
\newblock {OFA}: Unifying architectures, tasks, and modalities through a simple
  sequence-to-sequence learning framework.
\newblock In {\em International Conference on Machine Learning}, pages
  23318--23340. PMLR, 2022.

\bibitem{wu2018learning}
Jiajun Wu, Chengkai Zhang, Xiuming Zhang, Zhoutong Zhang, William~T Freeman,
  and Joshua~B Tenenbaum.
\newblock Learning shape priors for single-view 3d completion and
  reconstruction.
\newblock In {\em Proceedings of the European Conference on Computer Vision
  (ECCV)}, pages 646--662, 2018.

\bibitem{xie2018learning}
Jianwen Xie, Zilong Zheng, Ruiqi Gao, Wenguan Wang, Song-Chun Zhu, and
  Ying~Nian Wu.
\newblock Learning descriptor networks for 3d shape synthesis and analysis.
\newblock In {\em CVPR}, 2018.

\bibitem{xu2022discoscene}
Yinghao Xu, Menglei Chai, Zifan Shi, Sida Peng, Ivan Skorokhodov, Aliaksandr
  Siarohin, Ceyuan Yang, Yujun Shen, Hsin-Ying Lee, Bolei Zhou, et~al.
\newblock Discoscene: Spatially disentangled generative radiance fields for
  controllable 3d-aware scene synthesis.
\newblock In {\em CVPR}, 2023.

\bibitem{yang2015large}
Linjie Yang, Ping Luo, Chen Change~Loy, and Xiaoou Tang.
\newblock A large-scale car dataset for fine-grained categorization and
  verification.
\newblock In {\em Proceedings of the IEEE conference on computer vision and
  pattern recognition}, pages 3973--3981, 2015.

\bibitem{yu2021learning}
Rui Yu, Yue Dong, Pieter Peers, and Xin Tong.
\newblock Learning texture generators for 3d shape collections from internet
  photo sets.
\newblock In {\em BMVC}, 2021.

\bibitem{zeng2022lion}
Xiaohui Zeng, Arash Vahdat, Francis Williams, Zan Gojcic, Or Litany, Sanja
  Fidler, and Karsten Kreis.
\newblock Lion: Latent point diffusion models for 3d shape generation.
\newblock {\em arXiv preprint arXiv:2210.06978}, 2022.

\bibitem{zhang2023adding}
Lvmin Zhang and Maneesh Agrawala.
\newblock Adding conditional control to text-to-image diffusion models.
\newblock {\em arXiv preprint arXiv:2302.05543}, 2023.

\bibitem{zhang2021sketch2model}
Song-Hai Zhang, Yuan-Chen Guo, and Qing-Wen Gu.
\newblock Sketch2model: View-aware 3d modeling from single free-hand sketches.
\newblock In {\em CVPR}, 2021.

\end{thebibliography}
}

\clearpage
\appendix

\section*{Supplementary Material}

% \counterwithin{figure}{section}
% \counterwithin{table}{section}
% \counterwithin{equation}{section}

In this supplementary material, we provide the categories of the Objaverse subset in Sec.~\ref{sec:objaverse_subset} and the details of user study in Sec.~\ref{sec:user_study}.
To showcase the effectiveness of the proposed texture synthesis method, we provide additional results and analysis in Sec.~\ref{sec:additional_results}.

\begin{figure*}[!ht]
    \centering
    \includegraphics[width=0.99\linewidth]{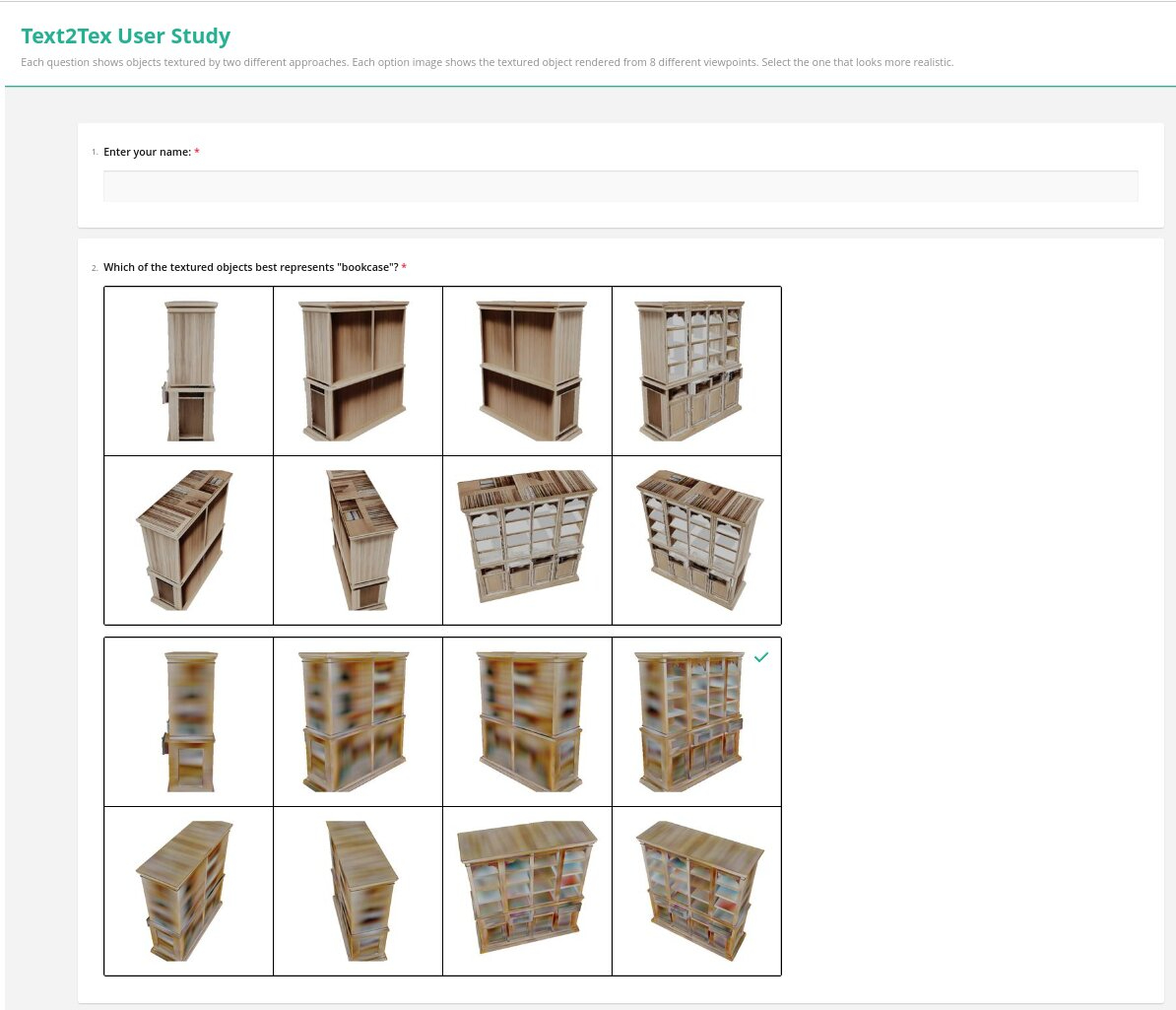}
    \caption{ 
    Screenshot of the user study interface.
    }
    \label{fig:user_study}
\end{figure*}

\begin{figure*}[!ht]
    \centering
    \includegraphics[width=0.8\linewidth]{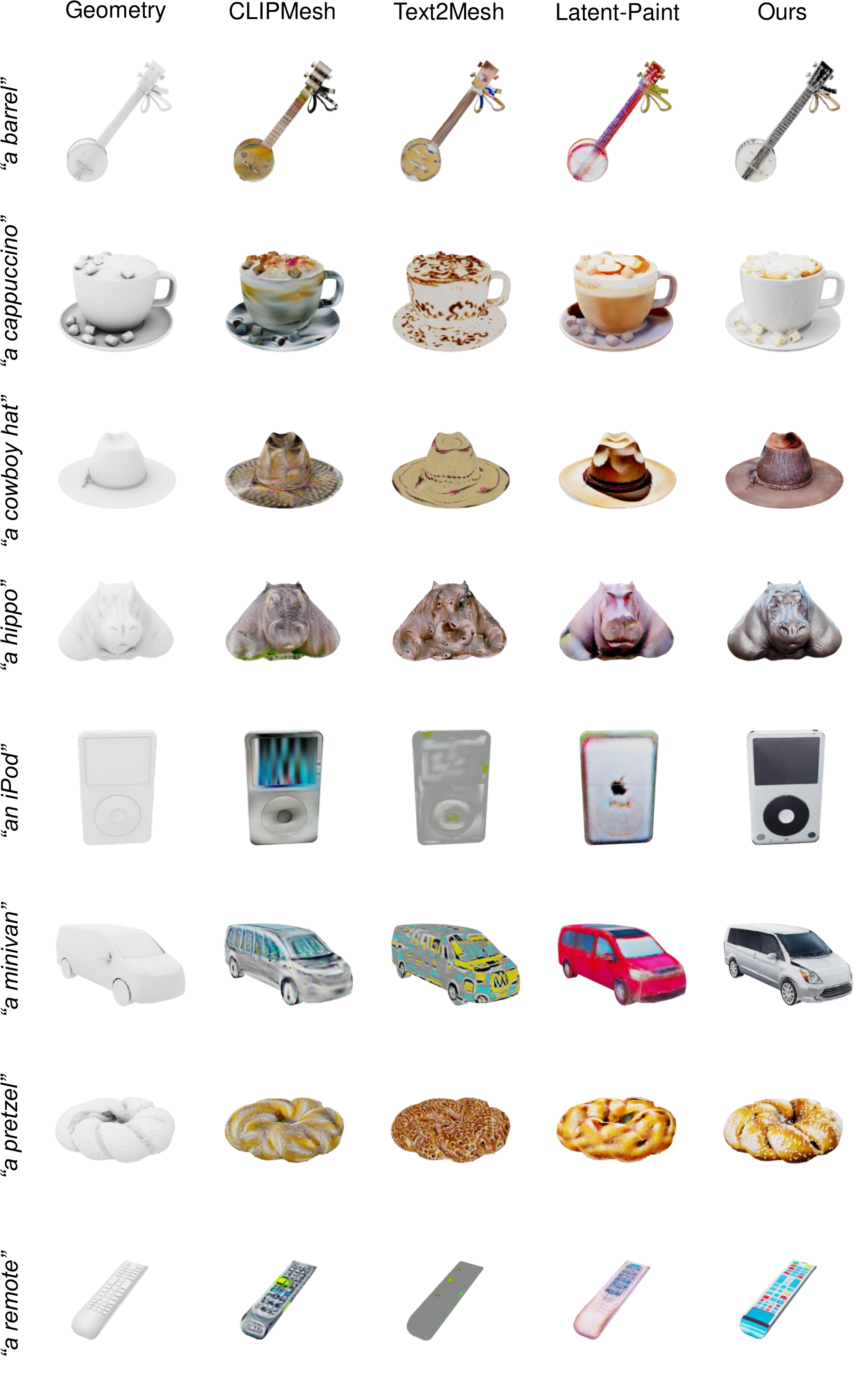}
    \caption{ 
    Additional qualitative comparison on objects from Objaverse~\cite{objaverse} dataset.
    }
    \label{fig:additional_results}
\end{figure*}

\begin{figure*}[!ht]
    \centering
    \includegraphics[width=0.99\linewidth]{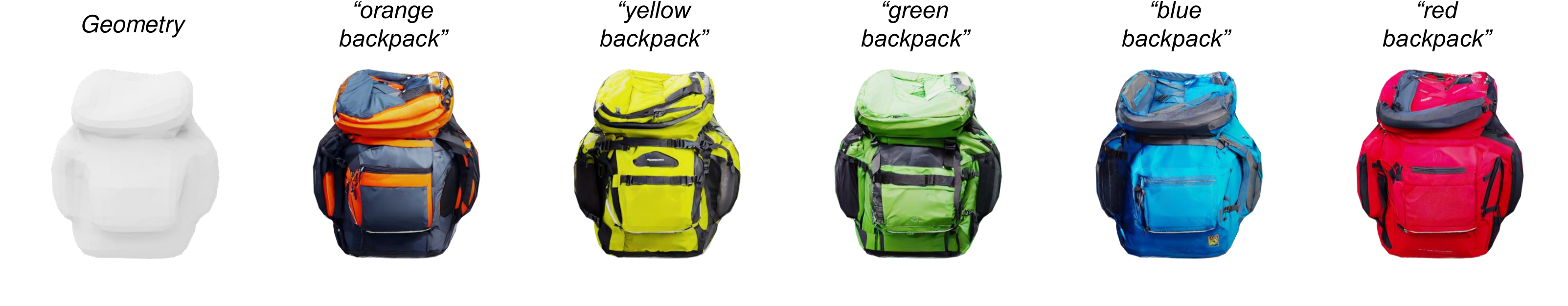}
    \caption{ 
    Different colors for the backpack. Our method loyally reflects the prompt colors in the textures.
    }
    \label{fig:different_color}
\end{figure*}

\begin{figure*}[!ht]
    \centering
    \includegraphics[width=0.99\linewidth]{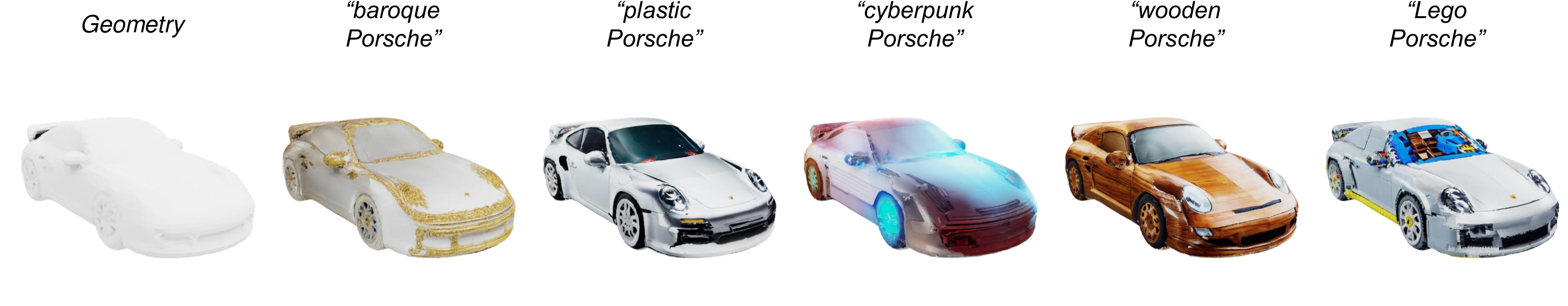}
    \caption{ 
    Different styles for the Porsche. Our method is capable of handling complicated styles such as ``baroque'' and ``cyberpunk'' without distorting the original properties of the input geometry.
    }
    \label{fig:different_style}
\end{figure*}

\begin{figure*}[!ht]
    \centering
    \includegraphics[width=0.99\linewidth]{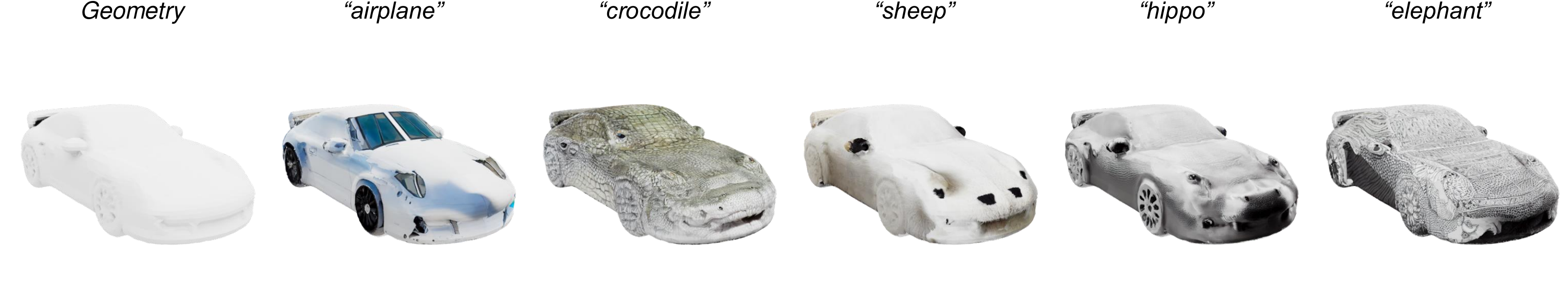}
    \caption{ 
    Creative textures for the Porsche with unrealistic prompts. Our method clearly represents the original properties of the geometry, while reflecting iconic characteristics of the input prompts.
    }
    \label{fig:wrong_prompt}
\end{figure*}

\section{Objaverse Subset}
\label{sec:objaverse_subset}

Our method is evaluated on a subset of the Objaverse~\cite{objaverse} dataset. To construct this subset, we first sample 3 random meshes from each category, where thin or unrecognizable meshes are filtered out. We also manually remove meshes with too simplistic textures and wrong categories. To reduce the processing time, we remove over-triangulated and scanned objects. After this curation, we obtain in total 410 high-quality textured objects across 225 categories. The chosen categories are as follows:

\begin{python}
# Objaverse subset object categories
["Bible", "CD player", "Lego", "Tabasco sauce", 
"aerosol can", "airplane", "alarm clock", 
"ambulance", "apple", "apricot", "armchair", 
"army tank", "baby buggy", "backpack", "bagel", 
"ball", "banana", "banjo", "barrel", "baseball", 
"baseball bat", "baseball glove", "basket", 
"basketball", "bathtub", "beanbag", "bed", 
"bedpan", "bench", "bicycle", "binoculars", 
"birdhouse", "birthday cake", "blimp", "boat", 
"bookcase", "bottle", "bowl", "bread", 
"briefcase", "broccoli", "broom", "bucket", 
"bulldog", "bulldozer", "burrito", "butterfly",
"cabinet", "calculator", "camera", "can", 
"candle", "canoe", "cappuccino", "carrot", 
"chair", "chaise longue", "chalice", 
"chocolate bar", "cigarette", "clipboard", 
"clock", "clutch bag", "coconut", "coffee maker",
"coffee table", "coffeepot", "comic book", 
"compass", "computer keyboard", "cowbell", 
"cowboy hat", "crate", "crown", "crucifix", 
"cucumber", "cup", "cupcake", "cushion", 
"dagger", "deck chair", "deer", "desk", "dog", 
"doll", "dolphin", "doughnut", "duffel bag", 
"dumbbell", "dumpster", "elephant", "fan", 
"faucet", "fighter jet", "file cabinet", 
"fire extinguisher", "first-aid kit", 
"fish", "flashlight", "forklift", "frog", 
"frying pan", "gargoyle", "giant panda", 
"globe", "glove", "goldfish", "goose", 
"guitar", "gun", "hair dryer", "hairbrush", 
"hamburger", "hammer", "hardback book", 
"heart", "helicopter", "helmet", "highchair", 
"hippopotamus", "hockey stick", "hourglass",
"hummingbird", "iPod", "jar", "jeep", 
"jet plane", "keg", "kettle", "key", "knife", 
"ladybug", "lantern", "laptop computer", 
"lemon", "lightbulb", "lizard", "machine gun", 
"martini", "matchbox", "microscope", 
"microwave oven", "milk", "milk can", "minivan", 
"money", "motorcycle", "muffin", "mug", 
"mushroom", "onion", "ottoman", "pancake", 
"pelican", "pen", "pencil", "pencil box", 
"piano", "pickup truck", "pie", "pigeon", 
"piggy bank", "pillow", "pistol", 
"pliers", "polar bear", "police cruiser", 
"pool table", "pot", "pretzel",
"pudding", "pumpkin", "race car", "radish", 
"rat", "refrigerator", "remote control", "rifle", 
"rocking chair", "saltshaker", "saucepan", 
"sausage", "school bus", "scissors", 
"screwdriver", "sewing machine", "shaker", 
"shark", "sheep", "shield", "shoe", 
"shopping bag", "shovel", "skateboard", 
"snowman", "soccer ball", "sofa", 
"sofa bed", "space shuttle", "spider", "stool", 
"suitcase", "sunflower", "sunglasses", "sunhat", 
"sushi", "sword", "syringe", "table", 
"table lamp", "teacup", "teakettle", "teapot", 
"teddy bear", "telephone", "television set",
"thermos bottle", "toilet", "toothbrush", 
"trailer truck", "trash can",
"tricycle", "truck", "typewriter", "umbrella", 
"urn","vending machine", "videotape", "violin", 
"watch", "water cooler", "water faucet", 
"watermelon", "wheel", "windmill", 
"wrench", "zucchini"]
\end{python}

\section{User Study Details}
\label{sec:user_study}

We develop a Django-based web application for the user study. In Fig.~\ref{fig:user_study}, we show the interface for the questionnaire. We randomly select 5 pairs of textured objects from each baseline and our method. To better visualize the samples, we render multi-view images for those objects from 8 preset viewpoints. After the samples are prepared, we ask the users to pick the sample from those pairs that best represents the text prompts. To avoid biases and cheating in this user study, we shuffle the pairs so that there is no positional hint of our method. In the end, we gather 604 responses from 41 participants to calculate the preferences.

\section{Additional Qualitative Results}
\label{sec:additional_results}

To further showcase the effectiveness of our method, we present additional qualitative results on objects from the Objaverse~\cite{objaverse} dataset.

\paragraph{Comparison with the baselines.}
We show additional comparisons against previous text-driven methods in Fig.~\ref{fig:additional_results}. In comparison with CLIP-based baselines (CLIPMesh~\cite{mohammad2022clip} and Text2Mesh~\cite{michel2022text2mesh}), our results are shown to be more detailed and realistic. Despite the blurry appearance, the results of Latent-Paint~\cite{metzer2022latent} still show competitive textures against ours. However, those results fail to capture the structural details of the input geometries. For instance, the scattered marshmallows in the case ``a cappuccino'' are incorrectly blended with the plate. In contrast, our method presents high-quality textures for the objects, while maintaining the correct structural details for the geometries.

\paragraph{Stylize the same objects.}
To show that our method can generate various textures for the same objects, we show different texturing results on the same objects. In Fig.~\ref{fig:different_color}, we show textures for the backpack with different colors in the prompts. All our textures are highly detailed and loyal to the colors, demonstrating the diversity and variety of the potential 3D contents. Moreover, we show that our method is not constrained by simple attributes. As shown in Fig.~\ref{fig:different_style}, our method is fully capable of reflecting complicated styles in the texture space, such as ``baroque'' and ``cyberpunk''. This indicates a great potential to stylize more high-quality 3D textures.

\paragraph{Creative texture synthesis.}
One interesting trait of our method is that it can move beyond certain categories. To show this, we showcase some creative textures on a Porsche in Fig.~\ref{fig:wrong_prompt}. Given some unrealistic prompts as input, our method is able to properly wrap the appearance on the geometry. For instance, in the case ``hippo'', our method aligns the eyes of a hippopotamus to the lamps of the Porsche as they are semantically similar to each other. It is worth mentioning that all textured objects clearly represents the original properties of the geometry (see the wheels of the case ``airplane''), while reflecting iconic characteristics of the input prompts (see the crocodile skin of the case ``crocodile'').

\end{document}